\documentclass[conference,usletter]{IEEEtran}
\usepackage{times,amsmath,amssymb}
\usepackage{slashbox}
\usepackage{graphicx}
\usepackage{xcolor}
\usepackage{algorithm}
\usepackage{algorithmicx}
\usepackage{setspace}
\usepackage[export]{adjustbox}
\usepackage{multirow}
\usepackage[noend]{algpseudocode}

\usepackage{mathabx}

\usepackage{framed} 
\immediate\write18{%
	makeindex -s nomencl.ist -o \jobname.nls -t \jobname.nlg \jobname.nlo%
}
\DeclareMathOperator*{\argmin}{arg\,min}

\usepackage{nomencl} 
\makenomenclature
\begin{document}

\title{Progressive Operational Perceptron with Memory}
\author{\IEEEauthorblockN{Dat Thanh Tran\IEEEauthorrefmark{1}, Serkan Kiranyaz\IEEEauthorrefmark{2}, Moncef Gabbouj\IEEEauthorrefmark{1}, Alexandros Iosifidis\IEEEauthorrefmark{3}}
\IEEEauthorblockA{\IEEEauthorrefmark{1}Department of Computing Sciences, Tampere University, Finland\\
\IEEEauthorrefmark{2}Department of Electrical Engineering, Qatar University, Qatar\\
\IEEEauthorrefmark{3}Department of Engineering, Electrical \& Computer Engineering, Aarhus University, Aarhus, Denmark\\
Email:\{thanh.tran, moncef.gabbouj\}@tuni.fi, mkiranyaz@qu.edu.qa, alexandros.iosifidis@eng.au.dk}\\
}

\maketitle

\begin{abstract}
Generalized Operational Perceptron (GOP) was proposed to generalize the linear neuron model used in the traditional Multilayer Perceptron (MLP) by mimicking the synaptic connections of biological neurons showing nonlinear neurochemical behaviours. Previously, Progressive Operational Perceptron (POP) was proposed to train a multilayer network of GOPs which is formed layer-wise in a progressive manner. While achieving superior learning performance over other types of networks, POP has a high computational complexity. In this work, we propose POPfast, an improved variant of POP that signiﬁcantly reduces the computational complexity of POP, thus accelerating the training time of GOP networks. In addition, we also propose major architectural modiﬁcations of POPfast that can augment the progressive learning process of POP by incorporating an information preserving, linear projection path from the input to the output layer at each progressive step. The proposed extensions can be interpreted as a mechanism that provides direct information extracted from the previously learned layers to the network, hence the term ``memory". This allows the network to learn deeper architectures and better data representations. An extensive set of experiments in human action, object, facial identity and scene recognition problems demonstrates that the proposed algorithms can train GOP networks much faster than POPs while achieving better performance compared to original POPs and other related algorithms.
\end{abstract}

\section{Introduction}\label{S:Intro}

Given a data set, a learning problem can be translated as the task of searching for the suitable transformation or mapping of the input data to some domains with specific characteristics. In discriminative learning, data in the target domain should be separable among different classes of input while in generative learning, data in the target domain should match some specific characteristics (e.g. a given distribution). In the biological learning system of mammals, the transformation is done by a set of neurons, each of which conducts electrical signals over three distinct operations: modification of the input signal from the synapse connection in the Dendrites; pooling operation of the modified input signals in the Soma, and sending pulses when the pooled potentials exceed a limit in the Axon hillock \cite{kiranyaz2017progressive}. Biological learning systems are generally built from a diverse set of neurons which perform various neuronal activities. For example, it has been shown that there are approximately $55$ different types of neurons to perform low-level visual sensing in mammalian retina \cite{masland2001neuronal}. 

\begin{figure}[t!]
	\centering
	\includegraphics[width=0.95\linewidth]{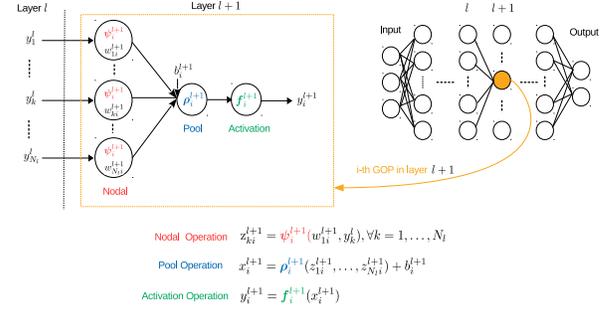}
	\caption{Activities of the $i$-th GOP neuron at layer $l+1$, characterized by the synaptic weights $w^{l+1}_{ki}$, the nodal operator $\boldsymbol{\psi}^{l+1}_i$, the pooling operator $\boldsymbol{\rho}^{k+1}_{i}$ and the activation operator $\boldsymbol{f}^{l+1}_i$}
	\label{f1}
\end{figure} 

In order to solve learning problems with machines, Artificial Neural Networks (ANNs) were designed to simulate biological learning system with artificial neurons as the core component. The most typical neuron model is based on McCulloch-Pitts perceptron \cite{mcculloch1943logical}, thereupon simply referred to as perceptron, which loosely mimics the behavior of biological neurons by scaling the input signals, summing over all scaled inputs, followed by the thresholding step. Mathematically, the activity of a perceptron corresponds to a linear transformation followed by an element-wise nonlinear function. Despite its simplicity, most of the existing state-of-the-art architectures in different application domains \cite{liu2017survey, guo2016deep, bajo2018neural, lipton2015critical} rely on this additive/affine perceptron model. This is due to the fact that linear transformation is expressed via matrix multiplication, which has several highly optimized implementations. While being efficient in terms of computation, the traditional perceptron model might not be optimal in terms of representation. In fact, the idea of enhancing the expressiveness of neural networks via more complex neuron models or activation functions has gradually attracted more attentions \cite{zhou2019functional, qian2018adaptive, jiang2018deep, fan2018universal}. In order to better simulate biological neuron in the mammalian nervous system, the authors in \cite{kiranyaz2017progressive} proposed a generalized perceptron model, known as Generalized Operational Perceptron (GOP), which admits a broader range of neuronal activities by three distinct sets of operations: nodal, pooling and activation operations. The schematic operation of GOP is illustrated in Figure \ref{f1}. 

As shown in Figure \ref{f1}, a GOP first applies a nodal operator ($\boldsymbol{\psi}^{l+1}_i$) to each individual output signal from the previous layer using adjustable synaptic weights $w^{l+1}_{ki}$ ($k=1,\dots, N_l$). The operated output signals are pooled to a scalar by the pooling operator ($\boldsymbol{\rho}^{l+1}_i$), after which the bias term $b_i^{l+1}$ is added. The activation operator ($\boldsymbol{f}^{l+1}_i$) determines the magnitude of activating signal that GOP sends to the next layer. By having the ability to select different nodal, pooling and activation operators from a library of operators, each GOP encapsulates a wide range of neural activities. For example, the traditional perceptron can be formed by selecting \textit{multiplication} as the nodal operator, \textit{summation} as the pooling operator and \textit{sigmoid} or \textit{ReLU} as the activation operator. In our work, the term \textit{operator set}, which refers to one specific choice of nodal, pooling and activation operator, represents a particular neuronal activity of a GOP. A sample library of operators is shown in Table \ref{t1}. Mathematically, the activities performed by the $i$-th GOP in layer $l+1$ can be described the by following equations:

\begin{align} 
z^{l+1}_{ki} &=  \boldsymbol{\psi}^{l+1}_{i}(y^{l}_{k}, w^{l+1}_{ki})\label{eq1} \\ 
x^{l+1}_{i} &= \boldsymbol{\rho}^{l+1}_{i}(z^{l+1}_{1i}, \dots, z^{l+1}_{N_{l}i}) + b^{l+1}_i\label{eq2} \\
y^{l+1}_{i} &= \boldsymbol{f}^{l+1}_{i}(x^{l+1}_i)\label{eq3}
\end{align} 

\begin{table}[]
	\begin{center}
		\caption{Operator set library}\label{t1}
		\resizebox{0.5\linewidth}{!}{
			\begin{tabular}{|c|c|}\hline
				
				\textbf{Nodal} ($\mathbf{\Psi}$)  		& \textbf{$\boldsymbol{\psi}^{l+1}_{i}(y^{l}_{k}, w^{l+1}_{ki})$} \\ \hline \hline
				Multiplication			  		& $w_{ki}^{l+1} y_k^l$ \\ \hline
				Exponential						& $\exp (w_{ki}^{l+1}y_k^l)-1$ \\ \hline
				Harmonic						& $\sin(w_{ki}^{l+1}y_k^l)$ \\ \hline
				Quadratic						& $w_{ki}^{l+1}(y_k^l)^2$ \\ \hline
				Gaussian						& $w_{ki}^{l+1} \exp(-w_{ki}^{l+1}(y_k^l)^2)$ \\ \hline
				DoG								& $w_{ki}^{l+1}y_k^l \exp(-w_{ki}^{l+1}(y_k^l)^2)$ \\ \hline \hline
				
				\textbf{Pool} ($\mathbf{P}$)		& $\boldsymbol{\rho}^{l+1}_{i}(z^{l+1}_{1i}, \dots, z^{l+1}_{N_{l}i})$ \\ \hline
				Summation 						& $\sum_{k=1}^{N_l} z_{ki}^{l+1}$ \\ \hline
				1-Correlation 					& $\sum_{k=1}^{N_l-1} z_{ki}^{l+1}z_{(k+1)i}^{l+1}$ \\ \hline
				2-Correlation 					& $\sum_{k=1}^{N_l-2} z_{ki}^{l+1}z_{(k+1)i}^{l+1}z_{(k+2)i}^{l+1}$ \\ \hline
				Maximum 						& $\underset{k}\max (z_{ki}^{l+1})$ \\ \hline \hline
				
				\textbf{Activation} ($\mathbf{F}$)		& $\boldsymbol{f}^{l+1}_{i}(x^{l+1}_i)$ \\ \hline
				Sigmoid					& $1 / (1+\exp(-x^{l+1}_i))$	\\ \hline
				Tanh					& $\sinh(x^{l+1}_i) /\cosh(x^{l+1}_i)$ \\ \hline
				ReLU					& $\max(0,x^{l+1}_i)$ \\ \hline
				
			\end{tabular}
		}
	\end{center}
\end{table}

Multiple GOPs can be combined to form multilayer network, hereafter called GOP networks. Since each GOP involves a library of operators, training a GOP network poses a much more challenging problem compared to standard MLP networks: not only the synaptic weights and the biases should be optimized but also the choice of the operator set per neuron. In \cite{kiranyaz2017progressive}, the authors proposed Progressive Operational Perceptron (POP), a specific configuration of GOP network in which each layer is progressively trained, given a pre-defined network template. To make the search of operator set tractable, POP constrains all GOPs within the same layer to share the same operator set, and the evaluation of each operator set is performed through stochastic optimization, i.e., Back Propagation (BP) algorithm. Recently, the authors in \cite{tran2018heterogeneous} proposed a new learning algorithm that aims at efficiency and compactness by constructing heterogeneous multilayer of GOPs utilizing a randomization process during the search procedure. 

In this study, we aim to improve the performance of POPs by making several modifications. Particularly, we incorporate a linear output layer relaxation to reduce the training complexity that only requires one iteration over the library of operator sets instead of four as in the original POP trained with two-pass GIS algorithm. In addition, we propose two memory schemes that aim to augment the progressive learning procedure in POP by incorporating an additional linear path that preserves information extracted from previous layers. The contributions of our work can be summarized as follows:

\begin{itemize}
	\item We propose POPfast, a simplified version of POP, which only requires one iteration over the library of operator sets compared to four iterations as in POP. Our experimental results demonstrate that POPfast performs similarly to POP while being faster.
	\item Based on POPfast, we propose two memory schemes to enable the network direct access to previous layers' information at each progressive step. For each memory scheme, we evaluate two types of information-preserving linear transformations to extract information synthesized by the previous layers. Extensive experiments were conducted to demonstrate performance improvements of POPfast augmented with memory. Besides, the importance of memory path is also empirically analyzed.  
	\item We make our implementation of all evaluated algorithms publicly available to facilitate future research, including parallel implementation for both single and multiple machines \cite{tran2019pygop}.
\end{itemize}

The remaining of the paper is organized as follows: In Section 2, we review POP and other related progressive algorithms for ANN training. Section 3 starts with the description of POPfast and continues to the description of the proposed memory schemes. In Section 4, we describe the details of our experimental setup, followed by quantitative analysis of the experiment results. Finally, our conclusion is made in Section 5. 

\begin{table*}[h!]
	\begin{framed}
		\nomenclature{GOP}{Generalized Operational Perceptron}
		\nomenclature{MLP}{Multilayer Perceptron}
		\nomenclature{POP}{Progressive Operational Perceptron}
		\nomenclature{POPfast}{Faster variant of POP}
		\nomenclature{POPmem-H}{Memory variant of POPfast with memory for hidden layer}
		\nomenclature{POPmem-O}{Memory variant of POPfast with memory for hidden \& output layer}
		\nomenclature{GIS}{Greedy Iterative Search}
		\nomenclature{MSE}{Mean Square Error}
		\nomenclature{SHLN}{Single Hidden Layer Network}
		\nomenclature{BP}{Back Propagation}
		\nomenclature{HeMLGOP}{Heterogeneous Multilayer Generalized Operational Perceptron algorithm}
		\nomenclature{BLS}{Broad Learning System algorithm}
		\nomenclature{S-ELM}{Stacked Extreme Learning Machine algorithm}
		\nomenclature{PLN}{Progressive Learning Network algorithm}
		\nomenclature{PCA}{Principal Component Analysis}
		\nomenclature{LDA}{Linear Discriminant Analysis}
		\nomenclature{$\mathcal{F}_l$}{Transformation performed by $l$-th GOP layer}
		\nomenclature{$\mathbf{X}_l$}{Input to the $l$-th hidden layer}
		\nomenclature{$\mathcal{G}_l$}{Information-preserving transformation in $l$-th hidden layer}
		\renewcommand{\nomname}{Nomenclature \& Abbreviation}
		\printnomenclature
	\end{framed}
\end{table*}

\section{Related Work}

This section reviews Progressive Operational Perceptron (POP) that is a particular type of GOP networks with progressive formation. In addition, other related progressive learning algorithms which were evaluated in our work are also briefly presented. 

\subsection{Progressive Operational Perceptron (POP)}

Given a target Mean Square Error (MSE) value and a network template $T=[I,h_1, \dots, h_N, O]$ that defines the number of hidden layers ($N$) and the number of neurons in each layer ($h_1, \dots, h_N$), POP sequentially learns one hidden layer at each step and terminates when the target MSE is achieved, or all layers in the template are learned. At step $k$, POP constructs a Single Hidden Layer Network (SHLN) with $h_{k-1}$ input neurons, $h_k$ hidden GOPs and $O$ output GOPs. With the constraint that neurons in the same layer share the same operator set, the learning task at step $k$ is to find the operator sets of the hidden and output layer with the synaptic weights that achieve the minimum MSE. This is done via a greedy iterative search procedure called two-pass GIS. 

Let $\phi_h$ and $\phi_o$ denote the operator set in the hidden and output layer respectively. In the first pass, $\phi_h$ is chosen randomly and fixed. The best performing $\phi_o^{*}$ is selected by iterating through all operator sets in the library and training the SHLN with $E$ epochs using Back Propagation (BP) algorithm at each iteration. Once $\phi_o^{*}$ is found, the algorithm continues by fixing $\phi_o^{*}$ and iterating through the library to find the best performing $\phi_h^{*}$. The second pass of GIS is similar to the first pass with the only exception that $\phi_h^{*}$ from the first pass is assigned to the hidden layer instead of a random assignment. The illustration of two-pass GIS algorithm is shown in the \ref{A1}, and the pseudo-code of POP is presented in Algorithm \ref{algo-pop}

After two-pass GIS, $\phi_h^{*}$ and the learned synaptic weights are fixed for the $k$-th hidden layer. If the MSE achieved by the current $k$ hidden layer network does not match the target MSE, POP discards the current output layer and continues to learn $(k+1)$-th hidden layer in the same manner. After the progression, if the target MSE value is not reached, POP fixes all the operator set assignments and finetunes all synaptic weights for some epochs. To learn a new hidden layer, it is clear that POP iterates four times over the library of operator set, requiring a complexity of $4N_OE$ BP epochs with $N_O$ is the total number of operator sets in the library.  

\begin{algorithm}
	\caption{Progressive Operational Perceptron (POP)}
	\label{algo-pop}
	\begin{algorithmic}[1]
		\State \textbf{Inputs}: 
		\State Training data $(\mathbf{X}, \mathbf{Y})$
		\State Network template $T = [I, h_1, \dots, h_N, O]$
		\State Target MSE threshold $\epsilon$
		\State The number of BP epoch $E$
		\State Library of operator sets $\mathbf{L} = \mathbf{\Psi} \bigtimes \mathbf{P} \bigtimes \mathbf{F}$
		\State \textbf{Training}:
		\State \textbf{for} $k \leftarrow 1$ to $N$ \textbf{do}
		\State \hskip1.5em Assign $\phi_{h_k}^*$ randomly from $\mathbf{L}$
		\State \hskip1.5em \textbf{repeat} twice: 
		\State \hskip3.0em \textbf{for} $\phi_i \in \mathbf{L}$ \textbf{do}
		\State \hskip4.5em Let $\phi_{h} = \phi_{h_k}^*$
		\State \hskip4.5em Let $\phi_O = \phi_i$
		\State \hskip4.5em Construct SHLN with:
		\State \hskip6.0em $h_k$ hidden GOPs having operator set $\phi_h$
		\State \hskip6.0em $O$ output GOPs having operator set $\phi_O$
		\State \hskip4.5em Optimize SHLN parameters $W_h^i$, $W_O^i$ for $E$ epochs
		\State \hskip4.5em Record loss value $l_i$
		\State \hskip3.0em Find $\displaystyle j = \argmin_{i} \{l_i\}$
		\State \hskip3.0em Assign $\phi_{O}^* = \phi_j$, $W_{h_k} = W_h^j$, $W_O = W_O^j$
		\State \hskip3.0em \textbf{for} $\phi_i \in \mathbf{L}$ \textbf{do}
		\State \hskip4.5em Let $\phi_O = \phi_O^*$
		\State \hskip4.5em Let $\phi_h = \phi_i$
		\State \hskip4.5em Construct SHLN with $\phi_h$, $\phi_O$
		\State \hskip4.5em Optimize SHLN parameters $W_h^i$, $W_O^i$ for $E$ epochs
		\State \hskip4.5em Record loss value $l_i$
		\State \hskip3.0em Find $\displaystyle j = \argmin_{i} \{l_i\}$
		\State \hskip3.0em Assign $l_k^* = l_j$, $\phi_{h_k}^* = \phi_j$, $W_{h_k} = W_h^j$, $W_O = W_O^j$
		\State \hskip1.5em \textbf{if} $l_k^* < \epsilon$ \textbf{then}
		\State \hskip3.0em \textbf{break}
		\State \textbf{Outputs}:
		\State $k$-hidden-layer network with GOPs' parameters $\phi_{h_i}^*$, $W_{h_i}$ ($i=1, \dots, k$), $\phi_O$, $W_O$
		
	\end{algorithmic}
\end{algorithm}

\subsection{Other Progressive Learning Algorithms}

Heterogeneous Multilayer Generalized Operational Perceptron (HeMLGOP) \cite{tran2018heterogeneous, tran2019learning, tran2019data, tran2019knowledge} is another progressive learning algorithm that was proposed to learn a heterogeneous architecture of GOPs by using a randomization technique \cite{huang2006extreme} during the operator set evaluation. The objective of HeMLGOP is, however, different from our POP with memory extension in that HeMLGOP is designed to learn efficient but compact network topologies while we aim to facilitate the progression to learn deeper architectures. While the literature in GOP is scarce, there are many progressive learning algorithms proposed for multilayer perceptron. 

Broad Learning System (BLS) \cite{chen2018broad} was proposed to extend the idea of Random Vector Functional Link Neural Network \cite{pao1994learning} by incrementing random neurons of a two hidden layer network. The first hidden layer extracts features through random linear transformation followed by sigmoid activation. Similarly, the second hidden layer applies a random linear transformation and sigmoid activation to the output of the previous layer. The features synthesized by both hidden layers are concatenated and fed to a linear classifier. BLS comes with efficient incremental solutions for both hidden layers and can be seen as a representative for the class of incremental randomized networks that have fixed depth. 

In \cite{zhou2015stacked}, the authors proposed Stacked Extreme Learning Machine (S-ELM) that progressively stacks several ELMs in a serial manner. The motivation of S-ELM is to divide a very large ELM network into multiple, connected ELMs to make the computation tractable. At each progressive step, S-ELM concatenates newly generated random features from the original input and previously synthesized hidden features which are extracted via Principle Component Analysis (PCA). The concatenated features are used to learn a linear classifier via least square solution. By retaining previously synthesized hidden features and generating new random features from the input, it can be considered that S-ELM virtually learns very large ELM. 

Similar to S-ELM, Progressive Learning Network (PLN) \cite{chatterjee2017progressive} also utilizes random transformation and previously synthesized information during progression. Different from S-ELM, PLN concatenates newly generated random features from the previous hidden layer output and information generated by the previously learned output layer. The concatenated features are fed to a linear classifier, which is solved by a constrained but convex optimization problem. Additionally, in PLN, blocks of random features are added to the current hidden layer until the performance saturates and the algorithm forms a new hidden layer. In this aspect, PLN is similar to HeMLGOP. 

In general, BLS, S-ELM, and PLN share the same objective as our proposed algorithm, i.e., to learn large and deep network architectures to achieve the best performances without factoring the cost of inference. S-ELM and PLN are similar to our work in that both algorithms augment the progressive learning by reusing past information. The specific motivation and mechanism of each algorithm and ours are, however, different.  

\section{Proposed Algorithms}

In this section, we start by describing POPfast, an extension we propose to reduce the training complexity of POP trained with two-pass GIS. We continue by describing our motivation to propose memory extensions to POPfast. Two memory extensions are then described and discussed in detail. 

\subsection{POPfast}

At each progressive step, POP constructs SHLN with the hidden and output layer based on GOPs. This requires the algorithm to search for the operator set of the hidden layer in conjunction with the output layer. A brute-force approach which evaluates all possible combination of operator sets in the hidden and output layer would require $N_O^2$ experiments with $E$ epochs each. By two-pass GIS, POP evaluates $4N_O$ experiments, which is only a small portion of the total search space. We propose to relax the output layer as a linear layer with appropriate activation function, i.e., soft-max for a classification task and identity for a regression task. By using a linear output layer, we enforce the network to learn successive nonlinear transformations that can lead to a feature space in which classes are linearly separable. This extension of POP is termed POPfast with the pseudo-code presented in Algorithm \ref{algo-popfast}. 

By fixing the form of the output layer, POPfast only needs to search for the operator set in the hidden layer when solving the SHLN configuration. The total search space of POPfast is, thus, $N_O$ experiments when adding a new hidden layer, which is $4\times$ smaller than the actual search space of POP ($4N_O$). The relaxation not only allows POPfast to be faster than POP when learning a new hidden layer but also guarantees that POPfast iterates through the whole search space. POP, on the other hand, only evaluates a fraction of the total search space ($4N_O$ out of $N_O^2$ configurations). 

\begin{algorithm}
	\caption{Faster Progressive Operational Perceptron (POPfast)}
	\label{algo-popfast}
	\begin{algorithmic}[1]
		\State \textbf{Inputs}: 
		\State Training data $(\mathbf{X}, \mathbf{Y})$
		\State Network template $T = [I, h_1, \dots, h_N, O]$
		\State Output activation function $\mathbf{f}_O$
		\State Target loss threshold $\epsilon$
		\State The number of BP epoch $E$
		\State Library of operator sets $\mathbf{L} = \mathbf{\Psi} \bigtimes \mathbf{P} \bigtimes \mathbf{F}$
		\State \textbf{Training}:
		\State \textbf{for} $k \leftarrow 1$ to $N$ \textbf{do}
		\State \hskip1.5em \textbf{for} $\phi_i \in \mathbf{L}$ \textbf{do}
		\State \hskip3.0em Let $\phi_{h} = \phi_i$
		\State \hskip3.0em Construct SHLN with:
		\State \hskip4.5em $h_k$ hidden GOPs having operator set $\phi_h$
		\State \hskip4.5em $O$ linear outputs, activation function $\mathbf{f}_O$
		\State \hskip3.0em Optimize SHLN parameters $W_h^i$, $W_O^i$ for $E$ epochs
		\State \hskip3.0em Record loss value $l_i$
		\State \hskip1.5em Find $\displaystyle j = \argmin_{i} \{l_i\}$
		\State \hskip1.5em Assign $l_k^* = l_j$, $\phi_{h_k}^* = \phi_j$, $W_{h_k} = W_h^j$, $W_O = W_O^j$
		\State \hskip1.5em \textbf{if} $l_k^* < \epsilon$ \textbf{then}
		\State \hskip3.0em \textbf{break}
		\State \textbf{Outputs}:
		\State $k$-hidden-layer network with:
		\State \hskip1.5em GOPs' parameters: $\phi_{h_i}^*$, $W_{h_i}$ for $i=1, \dots, k$
		\State \hskip1.5em Output layer's parameters: $W_O$
		
	\end{algorithmic}
\end{algorithm}

\subsection{Motivation}

Let $\mathbf{X}_l$ be the input to the $l$-th hidden layer, with $\mathbf{X}_1 = \mathbf{X}$ the input data. In addition, let $\mathcal{F}_l$ be the transformation performed by the $l$-th hidden layer. In POP and POPfast, when learning hidden layer $l$, the hidden layer is optimized with respect to the data representation $\mathbf{X}_l$ which is $\mathcal{F}_{l-1}(\mathbf{X}_{l-1})$ and the output layer only observes $\mathcal{F}_l(\mathbf{X}_{l})$ to learn a decision function. That means that the hidden layer and the output layer of the current SHLN do not have direct access to all previously extracted representations $\mathcal{F}_k(\mathbf{X}_{k})$ with $k=1, \dots, l-1$. If the size of $l$-th layer is not big enough or the transformation performed by $\mathcal{F}_l$ fails to produce more meaningful features, e.g. in terms of data discrimination, as compared to $\mathcal{F}_{l-1}$, the progression will terminate. From this viewpoint, learning new hidden layer as in POP and POPfast does not augment what has been learned by the entire network so far, but it can be interpreted as an attempt to learn better $\mathcal{F}_l(\mathbf{X}_{l})$ compared to $\mathbf{X}_l$ by only observing $\mathbf{X}_{l}$. Therefore, at each progressive step $l$, we aim to achieve two features to improve the progression of POPfast:

\begin{itemize}
	\item Instead of only $\mathcal{F}_{l-1}(\mathbf{X}_{l-1})$, we aim to provide the new hidden layer with the direct information from all previously learned representations $\mathcal{F}_{k}(\mathbf{X}_k)$, with $k=1, \dots, l-1$.
	\item In addition, we aim to provide the output layer of SHLN with the direct information from all previously learned representations $\mathcal{F}_{k}(\mathbf{X}_k)$, with $k=1, \dots, l-1$. 
\end{itemize}

By achieving the aforementioned two features, learning new hidden layer can then be understood as trying to complement what has been learned by the entire network so far. In the next subsection, we will propose two memory extensions: POPmem-H and POPmem-O. POPmem-H, which denotes the scheme that provides memory to the hidden layer, maintains the first feature. On the other hand, POPmem-O, which provides memory to the output layer, possesses both features mentioned above.  

\subsection{POPmem-H \& POPmem-O}\label{proposed}

\begin{figure*}[t!]
	\centering
	\includegraphics[width=0.98\textwidth]{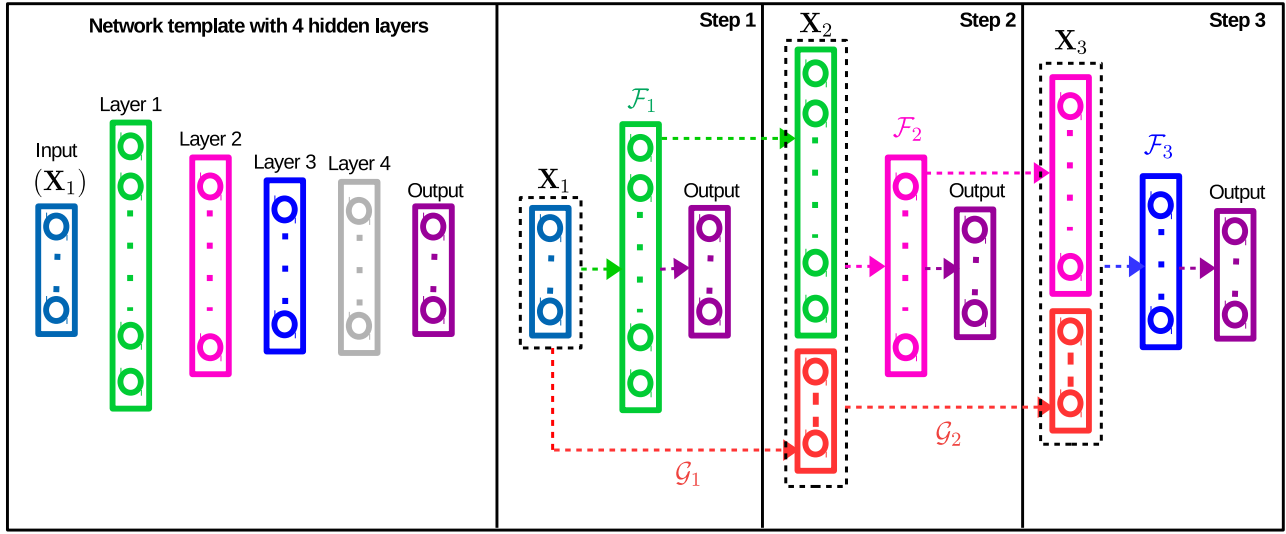}
	\caption{Progression in POPmem-H until the third hidden layer, given a 4-layer network template. At step $l$, POPmem-H forms a Single Hidden Layer Network (SHLN) with the input $\mathbf{X}_l$ formed by concatenating $\mathcal{F}_{l-1}(\mathbf{X}_{l-1})$ (the output of the previous GOP layer) and $\mathcal{G}_{l-1}(\mathbf{X}_{l-1})$ (the output of the linear transformation of $\mathbf{X}_{l-1}$). The linear transformation $\mathcal{G}_{l-1}$ is optimized using the respective algorithm, e.g., PCA or LDA, and then fixed. The hidden layer is a GOP layer ($\mathcal{F}_l$) which is optimized together with the output layer in a similar manner as POPfast. After that, $\mathcal{F}_l$ is fixed when POPmem-H proceeds to the next steps.}\label{f2}	
\end{figure*} 

Let $\mathcal{G}$ denotes a linear projection that preserves the information of the data. Depending on the form of $\mathcal{G}$, different types of information can be preserved. For example, Principal Component Analysis (PCA) tries to preserve the energy of the data, Linear Discriminant Analysis (LDA) aims to preserve the separability between different data classes, and Locality Preserving Projection (LPP) aims to preserve the structure of the local neighborhoods within the data. 

POPmem-H is similar to POPfast with the difference that at step $l$, instead of training the SHLN with $\mathcal{F}_{l-1}(\mathbf{X}_{l-1})$ as the input,  $\mathcal{F}_{l-1}(\mathbf{X}_{l-1})$ is concatenated with $\mathcal{G}_{l-1}(\mathbf{X}_{l-1})$ and the resulting vector is given as input to the SHLN. That is, at layer $l$, the input to the SHLN is $\mathbf{X}_l = [\mathcal{F}_{l-1}(\mathbf{X}_{l-1}), \mathcal{G}_{l-1}(\mathbf{X}_{l-1})]$. Thus, the new hidden layer is trained by observing information extracted from all previous layers. To better understand why $\mathbf{X}_l$ preserves information from all previous layers, we can see that $\mathcal{G}_{l-1}$ preserves information in $\mathbf{X}_{l-1}$, which is the concatenation of $\mathcal{F}_{l-2}(\mathbf{X}_{l-2})$ and $\mathcal{G}_{l-2}(\mathbf{X}_{l-2})$ and so on. We should note that $\mathcal{G}_l$, which is optimized based on its respective algorithm, e.g. generalized eigen-value decomposition for LDA, is fixed during the gradient descend updates of GOP neurons.

\begin{algorithm}
	\caption{Progressive Operational Perceptron with Augmented Memory for Hidden Layers (POPmem-H)}
	\label{algo-popmemh}
	\begin{algorithmic}[1]
		\State \textbf{Inputs}: 
		\State Training data $(\mathbf{X}, \mathbf{Y})$
		\State Network template $T = [I, h_1, \dots, h_N, O]$
		\State Output activation function $\mathbf{f}_O$
		\State Type of memory $\mathcal{G}$
		\State Target loss threshold $\epsilon$
		\State The number of BP epoch $E$
		\State Library of operator sets $\mathbf{L} = \mathbf{\Psi} \bigtimes \mathbf{P} \bigtimes \mathbf{F}$
		\State \textbf{Training}:
		\State Let $\mathcal{G}_0(\mathbf{X}_0) = \emptyset$
		\State Let $\mathcal{F}_0(\mathbf{X}_0) = \mathbf{X}$
		\State \textbf{for} $k \leftarrow 1$ to $N$ \textbf{do}
		\State \hskip1.5em Augment inputs with previous memory $\mathbf{X}_k = [\mathcal{F}_{k-1}(\mathbf{X}_{k-1}), \mathcal{G}_{k-1}(\mathbf{X}_{k-1})]$
		\State \hskip1.5em \# \textit{Find hidden GOPs ($\mathcal{F}_k$)}
		\State \hskip1.5em \textbf{for} $\phi_i \in \mathbf{L}$ \textbf{do}
		\State \hskip3.0em Let $\phi_{h} = \phi_i$
		\State \hskip3.0em Construct SHLN with:
		\State \hskip4.5em $\mathbf{X}_k$ as input data
		\State \hskip4.5em $h_k$ hidden GOPs ($\mathcal{F}_k$) having operator set $\phi_h$
		\State \hskip4.5em $O$ linear outputs, activation function $\mathbf{f}_O$
		\State \hskip3.0em Optimize SHLN parameters $W_h^i$, $W_O^i$ for $E$ epochs
		\State \hskip3.0em Record loss value $l_i$
		\State \hskip1.5em Find $\displaystyle j = \argmin_{i} \{l_i\}$
		\State \hskip1.5em Assign $l_k^* = l_j$, $\phi_{h_k}^* = \phi_j$, $W_{h_k} = W_h^j$, $W_O = W_O^j$
		\State \hskip1.5em \textbf{if} $l_k^* < \epsilon$ \textbf{then}
		\State \hskip3.0em \textbf{break}
		\State \hskip1.5em \# \textit{Find memory path ($\mathcal{G}_k$)}
		\State \hskip1.5em Solve for $W_{\mathcal{G}_k}$ that optimizes information in $\mathcal{G}_k(\mathbf{X}_k)$
		\State \textbf{Outputs}:
		\State $k$-hidden-layer network with:
		\State \hskip1.5em GOPs' parameters: $\phi_{h_i}^*$, $W_{h_i}$ for $i=1, \dots, k$
		\State \hskip1.5em Memory's parameters: $W_{\mathcal{G}_j}$ for $j=1,\dots, k-1$
		\State \hskip1.5em Output layer's parameters: $W_O$
		
	\end{algorithmic}
\end{algorithm}

\begin{figure*}[t!]
	\centering
	\includegraphics[width=0.98\textwidth]{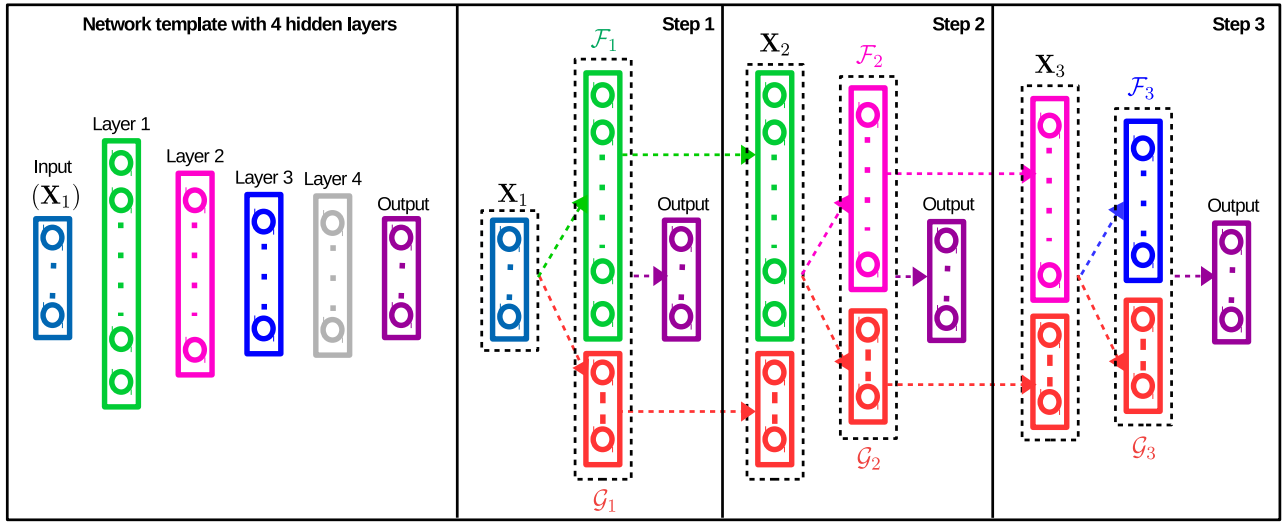}
	\caption{Progression in POPmem-O until the third hidden layer, given a network template of 4 hidden layers. At step $l$, POPmem-O forms a Single Hidden Layer Network (SHLN) with the input $\mathbf{X}_l$ formed by concatenating $\mathcal{F}_{l-1}(\mathbf{X}_{l-1})$ (the output of the previous GOP layer) and $\mathcal{G}_{l-1}(\mathbf{X}_{l-1})$ (the output of the linear transformation of $\mathbf{X}_{l-1}$). The hidden layer is the concatenation of the $l$-th GOP layer ($\mathcal{F}_l$) and the linear transformation $\mathcal{G}_l$. The linear transformation $\mathcal{G}_l$ is optimized with its respective algorithm, e.g., PCA or LDA, and POPmem-O learns $\mathcal{F}_l$ and the output layer in a similar manner as POPfast while fixing $\mathcal{G}_l$. After that, $\mathcal{F}_l$ is fixed when POPmem-O continues to the next steps.}
	\label{f3}
\end{figure*} 

While POPmem-H provides the new hidden layer with all previously synthesized hidden features, the output layer does not observe this information. If the new hidden layer cannot synthesize features as discriminative as the input, which preserves important information extracted from all previous layers, it is difficult for POPmem-H to learn a better output layer compared to the previous step. 

At step $l$, in order to provide both to the hidden and output layer with information related to the previously learned layers, we propose POPmem-O that incorporates the linear path $\mathcal{G}_l$ from the input to the output layer. This linear path is optimized with its respective algorithm and fixed before POPmem-O constructs the SHLN to learn new hidden layer $\mathcal{F}_l$. The optimization of $\mathcal{F}_l$ and the output layer in the SHLN is similar to POPfast. Since the hidden representation of SHLN at step $l$ is the concatenation of $\mathcal{F}_{l}(\mathbf{X}_l)$ and $\mathcal{G}_l(\mathbf{X}_l)$, $\mathbf{X}_{l+1} = [\mathcal{F}_{l}(\mathbf{X}_{l}), \mathcal{G}_{l}(\mathbf{X}_{l})]$ is used as the input to the $(l+1)$-th hidden layer. Therefore, the input to the hidden layer at each progressive step in POPmem-O contains all previously learned features by the network, similar to POPmem-H. Together with the linear path from the input to the output layer, it is obvious that POPmem-O achieves the two features that motivate us to exploit the augmented information in the progressive training process of POP. The pseudo-codes of POPmem-H and POPmem-O are presented in Algorithm \ref{algo-popmemh} and \ref{algo-popmemo}, respectively.  

\begin{algorithm}
	\caption{Progressive Operational Perceptron with Augmented Memory for Hidden \& Output Layers (POPmem-O)}
	\label{algo-popmemo}
	\begin{algorithmic}[1]
		\State \textbf{Inputs}: 
		\State Training data $(\mathbf{X}, \mathbf{Y})$
		\State Network template $T = [I, h_1, \dots, h_N, O]$
		\State Output activation function $\mathbf{f}_O$
		\State Type of memory $\mathcal{G}$
		\State Target loss threshold $\epsilon$
		\State The number of BP epoch $E$
		\State Library of operator sets $\mathbf{L} = \mathbf{\Psi} \bigtimes \mathbf{P} \bigtimes \mathbf{F}$
		\State \textbf{Training}:
		\State Let $\mathcal{G}_0(\mathbf{X}_0) = \emptyset$
		\State Let $\mathcal{F}_0(\mathbf{X}_0) = \mathbf{X}$
		\State \textbf{for} $k \leftarrow 1$ to $N$ \textbf{do}
		\State \hskip1.5em Augment inputs with previous memory $\mathbf{X}_k = [\mathcal{F}_{k-1}(\mathbf{X}_{k-1}), \mathcal{G}_{k-1}(\mathbf{X}_{k-1})]$
		\State \hskip1.5em \# \textit{Find memory path ($\mathcal{G}_k$)}
		\State \hskip1.5em Solve for $W_{\mathcal{G}_k}$ that optimizes information in $\mathcal{G}_k(\mathbf{X}_k)$
		\State \hskip1.5em \# \textit{Find hidden GOPs ($\mathcal{F}_k$)}
		\State \hskip1.5em \textbf{for} $\phi_i \in \mathbf{L}$ \textbf{do}
		\State \hskip3.0em Let $\phi_{h} = \phi_i$
		\State \hskip3.0em Construct SHLN with:
		\State \hskip4.5em $\mathbf{X}_k$ as input data
		\State \hskip4.5em Hidden layer with $h_k$ GOPs ($\phi_h$) and memory neurons $\mathcal{G}_k$
		\State \hskip4.5em $O$ linear outputs, activation function $\mathbf{f}_O$
		\State \hskip3.0em Optimize SHLN parameters $W_h^i$, $W_O^i$ for $E$ epochs, with $W_{\mathcal{G}_k}$ fixed
		\State \hskip3.0em Record loss value $l_i$
		\State \hskip1.5em Find $\displaystyle j = \argmin_{i} \{l_i\}$
		\State \hskip1.5em Assign $l_k^* = l_j$, $\phi_{h_k}^* = \phi_j$, $W_{h_k} = W_h^j$, $W_O = W_O^j$
		\State \hskip1.5em \textbf{if} $l_k^* < \epsilon$ \textbf{then}
		\State \hskip3.0em \textbf{break}
		\State \textbf{Outputs}:
		\State $k$-hidden-layer network with:
		\State \hskip1.5em GOPs' parameters: $\phi_{h_i}^*$, $W_{h_i}$ for $i=1, \dots, k$
		\State \hskip1.5em Memory's parameters: $W_{\mathcal{G}_j}$ for $j=1,\dots, k$
		\State \hskip1.5em Output layer's parameters: $W_O$
		
	\end{algorithmic}
\end{algorithm}

Figure \ref{f2} and \ref{f3} illustrate POPmem-H and POPmem-O when learning $l$-th hidden layer respective. Both memory schemes propose a generic way to augment the progressive learning procedure with an information-preserving linear projection $\mathcal{G}$. It should be noted that there exists other nonlinear transformations having similar properties such as Autoencoder \cite{ballard1987modular} or Variational Autoencoder \cite{kingma2013auto}. These methods, however, involve several hyper-parameters that require careful selection, which is done via extensive experiments. While having fewer hyper-parameters, LPP involves the eigenvalue decomposition of the kernel matrix, which scales badly as the number of training samples increases. By using only two simple dimensionality reduction methods, we are able to demonstrate the effectiveness of our proposed algorithms. While our memory extensions bear some resemblances to the skip-connection in ResNet or DenseNet \cite{he2016deep, huang2017densely}, there are certain differences: residual connection was proposed for static network architecture setting while our memory extensions are proposed for the progressive architecture learning setting with the motivation to learn new complementary hidden representation at each step; the memory extensions proposed in our work are in a generic form, allowing the adoption of any meaningful information preserving projection according to the problem at hand, which is optimized separately from the optimization of GOP hidden layers.  

\section{Experiments}

In this section, we detail our empirical evaluation and analysis of the proposed POPfast, POPmem-H and POPmem-O with respect to POP and three other related algorithms: BLS, S-ELM, and PLN. PCA and LDA were employed as the information-preserving, linear projection $\mathcal{G}$ in our memory proposals. The corresponding algorithms are denoted as POPmem-H-PCA, POPmem-H-LDA, POPmem-O-PCA, POPmem-O-LDA. 

Information related to the datasets, experimental protocol and implementation will be given first, followed by experimental results and discussion. The first set of experiments was conducted on small-scale datasets to demonstrate the efficacy of POPfast by having similar performance with reduced training complexity compared to POP. Since POP requires an enormous amount of computation on medium and large-scale datasets, the second set of experiments on those datasets was conducted without POP. 

\subsection{Datasets}

\begin{table}[t]
	\begin{center}
		\caption{Dataset Statistics}\label{t2}
		\resizebox{0.99\linewidth}{!}{
			\begin{tabular}{|c|c|c|c|}\hline
				
				Database 			& $\#$Samples			& Input dimension			& Target dimension 		\\ \hline \hline
				
				Olympic Sports \cite{niebles2010modeling}	& 774					& 100							& 16						\\ \hline
				Holywood3d \cite{hadfield2013hollywood}		& 945					& 100							& 14						\\ \hline \hline
				
				Caltech256 \cite{griffin2007caltech}		& 30607					& 512							& 257						\\ \hline	
				MIT indoor \cite{quattoni2009recognizing}		& 15620					& 512							& 67						\\ \hline	
				CFW60k \cite{zhang2012finding}				& 60000					& 512							& 500						\\ \hline		
				
			\end{tabular}
		}
	\end{center}
\end{table}

Our empirical evaluation contains results on $5$ classification problems of varying sizes: Olympic Sports \cite{niebles2010modeling}, Holywood3d \cite{hadfield2013hollywood}, Caltech256 \cite{griffin2007caltech}, MIT indoor \cite{quattoni2009recognizing} and CFW60k \cite{zhang2012finding}. Statistics about the datasets are shown in Table \ref{t2}. 

Olympic Sports and Holywood3d represents the problem of human action recognition in videos. Caltech256 is an object classification dataset with $256$ objects and one background class. MIT indoor is used for indoor scene recognition with $66$ different indoor scene categories. CFW60k, which is a subset of Celebrity in the Wild (CFW) dataset \cite{zhang2012finding}, contains $60$K facial images depicting $500$ celebrities. CFW60k was used as a face recognition dataset in our experiments. 

In order to extract meaningful video representation for Olympic Sports and Holywood3d, we adopted the state-of-the-art descriptor proposed in \cite{wang2013action} and combined five action descriptions using the suggested multi-channel kernel approach, with which Kernel PCA was applied to obtain $100$-dimensional vector-based representation for each video. Regarding Caltech256 and MIT indoor, deep features were extracted by average pooling over the spatial dimension of the last convolution layer of VGG network \cite{simonyan2014very} pre-trained on ILSVRC2012 database. Similar deep features were generated for CFW60k using VGGface network \cite{parkhi2015deep}.

\subsection{Experiment Protocol}

\begin{table}[t]
	\begin{center}
		\caption{Classification performance (\%) on small-scale datasets}\label{t3}
		\resizebox{0.85\linewidth}{!}{
			\begin{tabular}{|c|c|c|}\hline
				
				& Holywood3d			& Olympic Sports			\\ \hline \hline
				POP				& $78.03$				& $87.30$ \\ \hline
				POPfast			& $79.42$				& $87.49$ \\ \hline \hline
				POPmem-H-PCA	& $80.32$				& $88.31$ \\ \hline
				POPmem-H-LDA	& $78.36$				& $88.70$ \\ \hline \hline
				POPmem-O-PCA	& $\mathbf{80.65}$		& $\mathbf{88.71}$ \\ \hline
				POPmem-O-LDA	& $78.68$				& $87.90$ \\ \hline \hline
				S-ELM			& $72.78$				& $83.06$ \\ \hline
				BLS				& $73.77$				& $81.85$ \\ \hline
				PLN				& $72.13$				& $76.61$ \\ \hline
				
			\end{tabular}
		}
	\end{center}
\end{table}

For Olympic Sports and Holywood3d, the standard partition provided by the database was used in our experiments. With Caltech256, MIT indoor and CFW60k, we randomly shuffled and employed $60\%$ of the data for training and $20\%$ each for validation and testing. When the validation set is available, the performance measured on the validation set is used to determine the stopping criterion and the performance on the test set is reported in this paper with the median over three runs. 

While POP was originally proposed with an absolute measure of the stopping criterion, we applied a relative measure to determine when to stop the progression to every evaluated algorithm, which ensures a fair progression setting for all algorithms. Particularly, let $A_l$ denotes the accuracy achieved at the $l$ progressive step, the progression stops when

\begin{equation}
\frac{A_{l}-A_{l-1}}{A_{l-1}} < 10^{-4}
\end{equation} 

Regarding the regularization methods for GOP-based algorithms, $50\%$ of Dropout was applied to the output of the hidden layers. In addition, two types of weight regularization were experimented individually: weight decay and $l_2$ norm constraint. The coefficient for weight decay was set to $0.0001$ and the maximum norm value was set to $2.0$. During the operator set evaluation, each network was trained for $300$ epochs with the initial learning rate equal to $0.01$ that drops by $0.01$ after every $100$ epochs. After the progression, the entire network was finetuned for $200$ epochs with initial learning rate $0.0001$ that drops to $0.00001$ after $100$ epochs. A network template of $8$ hidden layers, each of which has $40$ GOPs, was given to all GOP-based algorithms. When PCA is employed as the memory path, the subspace dimension was selected as the minimum number of principal axes required to keep $98\%$ of the energy. In case of LDA, the subspace dimension was fixed to $C-1$ with $C$ is the number of target classes. For both projections, the data is centered at the origin and $0.01$ was added to the diagonal of the covariance matrix in case of singularity.

\begin{table}[t]
	\begin{center}
		\caption{Training time (second) per layer on small-scale datasets}\label{t4}
		\resizebox{0.85\linewidth}{!}{
			\begin{tabular}{|c|c|c|}\hline
				
				& Holywood3d			& Olympic Sports			\\ \hline \hline
				POP				& $48484$				& $28414$ \\ \hline
				POPfast			& $7851$				& $6881$ \\ \hline \hline
				POPmem-H-PCA	& $7921$				& $6990$ \\ \hline
				POPmem-H-LDA	& $8330$				& $7291$ \\ \hline \hline
				POPmem-O-PCA	& $10549$				& $8905$ \\ \hline
				POPmem-O-LDA	& $10507$				& $7804$ \\ \hline \hline
				S-ELM			& $11$					& $10$ \\ \hline
				BLS				& $3$					& $4$ 	\\ \hline
				PLN				& $178$					& $182$ \\ \hline
				
			\end{tabular}
		}
	\end{center}
\end{table}

Regarding BLS, S-ELM, and PLN, we have experimented with a wide range of hyper-parameters since these methods are sensitive to the hyper-parameter selection. For BLS, the regularization applied to pseudo-inverse ($\lambda$) and regularization coefficient used in Alternating Direction Method of Multiplier (ADMM) ($\mu$) was selected from the set $\{10^{-3}, 10^{-2}, 10^{-1}, 1, 10, 10^2, 10^3\}$. The same range was used in PLN for least-square regularization ($\lambda$), output layer optimization ($\alpha$ and $\mu$), and in S-ELM for least-square regularization. The number of iterations in ADMM was set to $500$ for both PLN and BLS. For S-ELM, we followed Algorithm 2 as given in \cite{zhou2015stacked} and concatenated $500$ new hidden neurons with $500$ hidden features extracted by PCA from the previous layer at each progressive step. In BLS and PLN, the incremental step is $20$ and the maximum number of random neurons per hidden layer was fixed to $1000$.

Table \ref{t3} shows the classification performance of all evaluated algorithms on two small datasets. In order to demonstrate the effectiveness of POPfast and memory extensions in terms of training time compared to POP, we conducted all algorithms on a single machine with the same configuration and report the training time per layer on two small-scale datasets in Table \ref{t4}. For medium and large-scale datasets, experiments were conducted on a cluster operating with a queuing system, thus the training times of different algorithms are not comparable and omitted here. It is clear that POPfast has similar or better performance compared to POP with relatively shorter training time per layer. Among all algorithms, POPmem-O-PCA is the best performing algorithm on both datasets while S-ELM, BLS, and PLN are inferior to GOP-based algorithms. While memory variants utilizing PCA consistently outperform POPfast, it is not the case with LDA. Since the memory extensions require an additional step to calculate the linear projection, the training time of POPmem-O and POPmem-H are slightly slower than POPfast but still far more efficient as compared to POP. Without involving the operator set searching step, perceptron-based algorithms, i.e. BLS, S-ELM, and PLN, are the fastest to train.

\begin{table}[t]
	\begin{center}
		\caption{Classification performance (\%) on medium and large-scale datasets}\label{t5}
		\resizebox{0.9\linewidth}{!}{
			\begin{tabular}{|c|c|c|c|}\hline
				
				& Caltech256			& MIT indoor				& CFW60K			\\ \hline \hline
				POPfast			& $73.93$				& $66.82$ 					& $85.05$			\\ \hline
				POPfast*		& $77.62$				& $68.37$ 					& $87.46$			\\ \hline \hline
				POPmem-H-PCA	& $74.43$				& $66.98$ 					& $84.61$			\\ \hline
				POPmem-H-LDA	& $74.04$				& $66.76$ 					& $84.79$ 			\\ \hline \hline
				POPmem-O-PCA	& $79.25$				& $\mathbf{69.04}$	 		& $\mathbf{88.95}$ 	\\ \hline
				POPmem-O-LDA	& $\mathbf{79.35}$		& $68.22$ 					& $88.89$			\\ \hline \hline
				S-ELM			& $69.83$				& $60.83$ 					& $64.40$			\\ \hline
				BLS				& $72.35$				& $58.35$ 					& $75.92$			\\ \hline
				PLN				& $75.57$				& $65.85$ 					& $85.79$ 			\\ \hline
				
			\end{tabular}
		}
	\end{center}
\end{table}

Since POP requires a large amount of computation, experiments on medium and large-scale datasets were not conducted for POP. The classification performances of all other algorithms are shown in Table \ref{t5}. It is obvious that both PCA and LDA variants of POPmem-H indicate no improvement as compared to POPfast. On the other hand, there are huge gaps between POPmem-O variants and POPfast or POPmem-H. The differences between two variants of POPmem-O are relatively small. As discussed in Section \ref{proposed}, during the progression in POPmem-H, information learned from all previous layers can be observed by the new hidden layer but not the output layer. Thus, POPmem-H might struggle to learn new hidden layer that synthesizes better features compared to all previously extracted features preserved in the input of the SHLN. On the contrary, the memory path in POPmem-O allows both hidden and output layer to access information related to previously learned layers, which augments the network to learn better representation. 

In order to empirically verify the importance of the linear memory path, we took the network topologies learned by POPmem-O as the templates to train POPfast and denote the results as POPfast*. While improving over POPfast due to larger hidden layers, the performances of POPfast* are still inferior to POPmem-O variants. This indicates that the hidden layers in POPmem-O composing of both nonlinear neurons (GOPs) and information-preserving linear neurons produce more discriminative representations compared to those in POPfast* with only GOPs. 

Since S-ELM and BLS utilize only random hidden neurons, they perform worse than other evaluated algorithms. As in case of PLN, the algorithm performs better than POPfast on Caltech256 and CFW60K but worse on MIT indoor. This is due to the fact that each hidden layer in PLN is formed by newly added neurons and features produced by the previous prediction, which is always twice the number of classes. That is, in Caltech256 and CFW60K, hidden layers of PLN have at least $502$ and $1000$ neurons respectively, which are much larger than dimensions in POPfast. When the network templates increase as in POPfast*, PLN is outperformed on all three datasets.

\section{Conclusions}

In this paper, we proposed POPfast, an efﬁcient algorithm that accelerates the training time of the original POP algorithm while achieving competitive performance in a variety of classification problems. Since learning with GOPs involves operator set evaluation, our work contributes an efﬁcient search procedure for the future works that employ GOPs, enabling us to tackle more complex and larger datasets as illustrated in our experiments. Based on the accelerated search procedure, we propose two architectural extensions, i.e. POPmem-H and POPmem-O algorithms, that aim to augment the progressive learning procedure by exploiting information learned from all previous layers. The memory variants of POPfast propose a novel approach that addresses the problem of “learning complementary representations” in progressive learning. This approach is applicable not only to GOP networks, but can also be used for other types of progressive learning models. Our empirical analysis shows that when complementary representation is explicitly learned at each progressive step as in POPmem-H and POPmem-O, the resulting networks learn better representations and outperform those generated by other progressive learning algorithms.

\appendix

\section{Two-pass GIS}\label{A1}

\begin{figure*}[h!]
	\centering
	\includegraphics[width=0.9\textwidth]{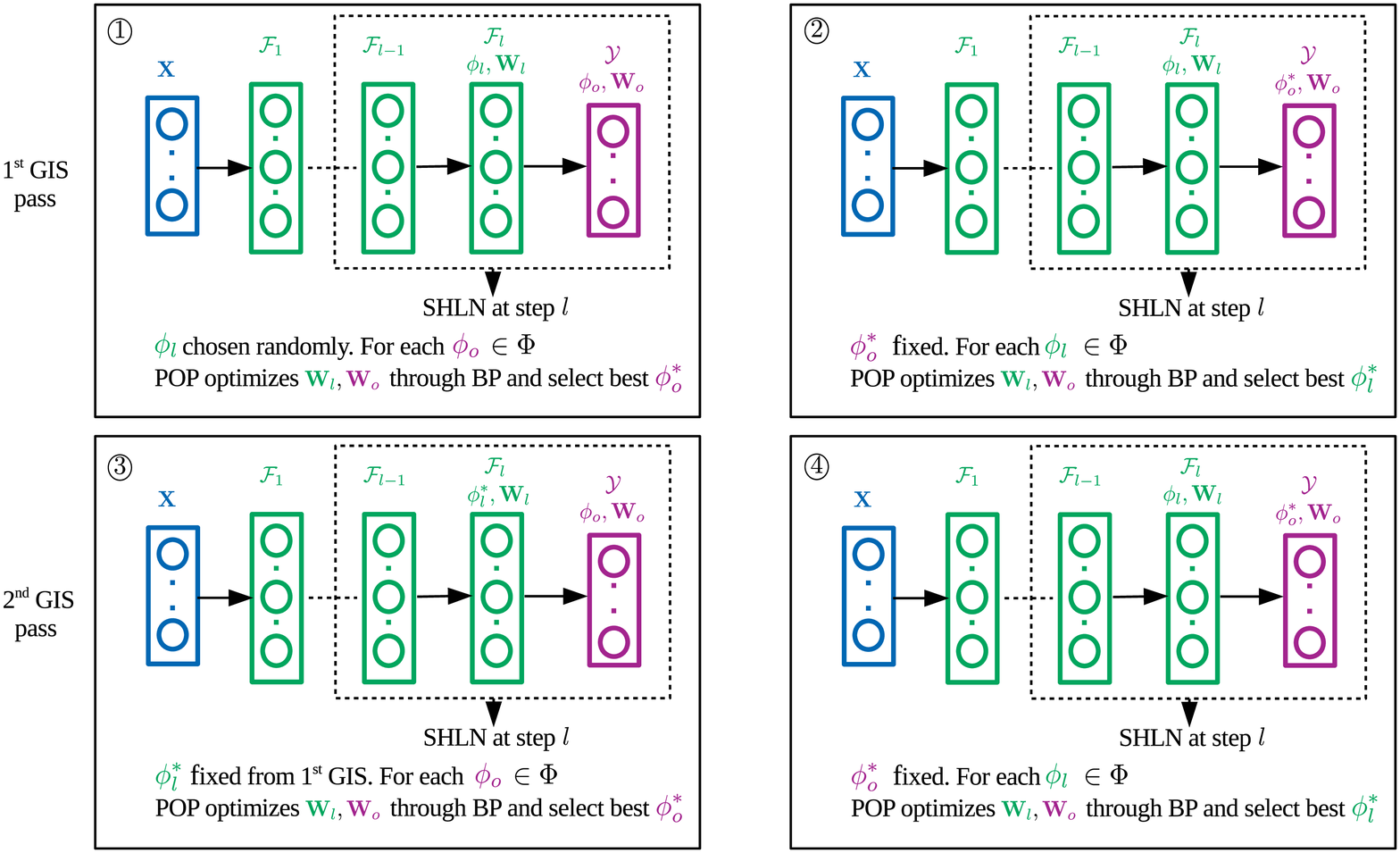}
	\caption{Four steps $\textcircled{\small{1}}, \textcircled{\small{2}}, \textcircled{\small{3}}, \textcircled{\small{4}}$ in two-pass GIS algorithm.}\label{f4}	
\end{figure*}

Here we illustrate the two-pass GIS algorithm to train a Single Hidden Layer Network (SHLN) in the original POP network. It starts by randomly selecting the operator set for the hidden layer as illustrated in step $\textcircled{\small{1}}$. Given this random initialization, the entire library of operator set is evaluated for selecting the operator set of the output layer. Given the selected operator set for the output layer, the entire operator set library is again evaluated for selecting the operator set of the hidden layer. These two steps are repeated in the second pass of GIS as illustrated in the bottom row of Figure \ref{f4}.

\bibliography{reference}
\bibliographystyle{ieeetr}

\end{document}